\documentclass[pdflatex,sn-mathphys-num]{sn-jnl}

\usepackage{algorithm}
\usepackage{algpseudocode}
\usepackage{graphicx}
\usepackage{tikz}
\usepackage{amsmath}
\usepackage{amssymb}
\usepackage{multirow}
\usepackage[section]{placeins}
\usepackage{lscape}
\usepackage{tabularx}
\usepackage{caption}
\usepackage{subcaption}
\usepackage{bbm}
\usepackage{acro}
\usepackage{tocloft}
\usepackage{url}

\PassOptionsToPackage{hidelinks}{hyperref}
\PassOptionsToPackage{hyphens}{url}
\usepackage{hyperref}

\usetikzlibrary{3d, decorations.pathreplacing}

\theoremstyle{thmstyleone}%
\theoremstyle{thmstyletwo}%

\theoremstyle{thmstylethree}%

\begin{document}

\title[AUEC Framework for Unsupervised Learning]{Autoencoded UMAP-Enhanced Clustering for Unsupervised Learning}


\author[1]{\fnm{Malihehsadat} \sur{Chavooshi}}\email{malihehsadat.chavooshi@bcm.edu}

\author*[2]{\fnm{Alexander} V. \sur{Mamonov}}\email{avmamonov@uh.edu}

\affil[1]{\orgdiv{Medicine - Health Services Research}, \orgname{Baylor College of Medicine}, \orgaddress{\street{One Baylor Plaza}, \city{Houston}, \state{TX}, \postcode{77030}, \country{USA}}}

\affil[2]{\orgdiv{Department of Mathematics}, \orgname{University of Houston}, \orgaddress{\street{3551 Cullen Blvd}, \city{Houston}, \state{TX}, \postcode{77204-3008}, \country{USA}}}


\abstract{We propose a novel approach to unsupervised learning by constructing a non-linear embedding of the data into a low-dimensional space followed by any conventional clustering algorithm. The embedding promotes clusterability of the data and is comprised of two mappings: the encoder of an autoencoder neural network and the output of UMAP algorithm. The autoencoder is trained with a composite loss function that incorporates both a conventional data reconstruction as a regularization component
and a clustering-promoting component built using the spectral graph theory. The two embeddings and the subsequent clustering are integrated into a three-stage unsupervised learning framework, referred to as Autoencoded UMAP-Enhanced Clustering (AUEC). When applied to MNIST data, AUEC significantly outperforms the state-of-the-art techniques in terms of clustering accuracy.}





\maketitle

\section{Introduction}
\label{sec:intro}

Clustering is a fundamental tool in unsupervised machine learning, data mining and pattern recognition. 
However, it remains notoriously challenging when the underlying topology of the data manifold is complicated.
Thus, the best clustering techniques are expected to include some form of manifold learning to gain an insight into the topological structure of the data before performing clustering itself. One possible approach to learn and simplify the topology of the data is to employ dimensionality reduction (DR) techniques prior to clustering. A large variety of such methods exists, ranging from basic ones such as the principal component analysis (PCA) to more advanced nonlinear DR
approaches like Uniform Manifold Approximation and Projection (UMAP) \cite{becht2019umap}, 
Laplacian eigenmaps \cite{belkin2001laplacian}, LargeVis \cite{tang2016largevis} and many more.

Deep neural networks (DNNs) \cite{schmidhuber2015deep} represent another class of techniques that is increasingly employed for DR before clustering. Examples include the stacked autoencoder (SAE) \cite{vincent2010stacked}, deep CCA (DCCA) \cite{andrew2013deep}, and sparse autoencoders \cite{ng2011sparse} that learn nonlinear mappings from the data domain to low-dimensional latent spaces. These approaches treat DNNs as a separate preprocessing stage from clustering, hoping that the latent representations learned will naturally be suitable for clustering. However, without explicitly incorporating a clustering-promoting objective in the learning process, the resulting DNNs do not necessarily produce reduced-dimension data amenable to clustering. Thus, a number of recent works has explored merging DR with clustering rather than using DR merely as a preprocessing tool, e.g., \cite{soete1994, patel2013joint, yang2017towards}. These are the so-called unified approaches that optimize both deep representation learning and clustering objectives simultaneously. All of them are build on the assumption of the existence of a latent space where entities form distinct clusters. Therefore, it is logical to look for a DR transformation that unveils this structure, e.g., one that results in a small K-means loss function. This led to the idea of using the K-means cost function in the latent space to guide DR toward producing data representations amenable to K-means clustering.

Following the above considerations, a number of approaches was developed. 
For example, the so-called Deep Embedded Clustering (DEC) \cite{xie2016} maps the observed space to a lower-dimensional latent space using SAE, simultaneously deriving feature representations and cluster assignments. Further improvements were later introduced in the form of Deep Clustering Network (DCN) \cite{li2018discriminatively} that augments DEC by substituting SAE with a convolutional autoencoder (CAE), while IDEC \cite{guo2017improved} integrates the reconstruction loss of autoencoders into DEC's objective \cite{xie2016}. In all the approaches discussed above, the clustering module is linked to DNN's output, aiming for a simultaneous learning of both DNN parameters and cluster assignments. This leads to an optimization problem 
\begin{equation}
\mathop{\mbox{minimize}}_{w,\theta} 
\sum_{i=1}^{N} q(f(x_i; w); \theta),
\label{eq:jointdeep}
\end{equation}
where \( f(x_i; w) \) is the output of the network given data instance \( x_i \), \( w \) contains the network weights, and \( \theta \) corresponds to parameters of a specific clustering model. For example, for K-means clustering 
\( \theta \) contains the centroids and cluster assignments. Here, \( q \) stands for a clustering loss, such as the Kullback–Leibler (KL) divergence loss 
featured in \cite{xie2016} or the agglomerative clustering loss from \cite{yang2016joint}, possibly with regularization terms added to it. 

While the approaches \cite{yang2016joint,xie2016,li2018discriminatively,guo2017improved} proved a certain degree of effectiveness, there is still room for improvement. First, some improvement may be achieved from using an alternative clustering loss. Here we employ the spectral graph theory to derive such a loss. Second, one may notice that using DNN alone may be insufficient for learning an efficient representation of the data and therefore may lead to sub-optimal clustering results. Thus, we propose to augment DNN-based DR with a secondary DR step that utilizes the power of Uniform Manifold Approximation and Projection (UMAP) \cite{mcinnes2018umap} to refine the DNN embedding and further improve the clusterability of the embedded data. This results in a three-stage unsupervised learning framework that we refer to as Autoencoded UMAP-Enhanced Clustering (AUEC).

The rest of the paper is organized as follows. We introduce the AUEC framework in Section~\ref{sec:auec} and give an overview of its three stages. Since the first stage involving DNN training with a composite loss function is the most complicated of all three, we dedicate Section~\ref{sec:compress} to discussing it in detail.
We evaluate the performance of AUEC on the MNIST dataset and compare the results to the state-of-the-art approaches in Section~\ref{sec:numerics}. Finally, conclusions are made in Section~\ref{sec:conclude}

\section{Autoencoded UMAP-Enhanced Clustering framework}
\label{sec:auec}

In this section, we introduce the AUEC framework for unsupervised machine learning that combines 
the strengths of AEs with a carefully designed joint loss to process the input data and UMAP to enhance 
the DR and subsequent clustering. It consists of three stages, as explained below.

\textbf{Stage I: Autoencoder with Joint loss}

The first stage of AUEC is DR performed by means of embedding the data $X = \{x_i\}_{i=1}^{N}$ into a latent space via an autoencoder-like DNN trained using a joint loss. We refer to this primary DR as the compressed embedding.
The joint loss is given by a weighted sum of two components. First, is the component that measures the quality of clustering (clusterability) of the data embedded into the latent space. We refer to this component as the clustering loss. It can employ a variety of different clusterability measures as discussed below. Its purpose is to enforce a topology of the latent space that emphasizes the cluster structure of the data.
Second, is the conventional reconstruction loss of an auto-encoder. Its presence provides a regularization effect on the network training by preventing trivial embeddings that may be produced by minimizing the clustering loss on its own. Thus, we refer to it as the regularization loss. The purpose of the first stage is to untangle the local cluster structure of the data. However, it may be insufficient to obtain the most optimal embedding for subsequent clustering. Therefore, it is followed by another DR stage.

\textbf{Stage II: Secondary DR via UMAP}

To further enhance clusterability in the latent space, we follow the compressed embedding by a secondary DR step performed via UMAP \cite{mcinnes2018umap}. It is powerful nonlinear DR technique that combines the ideas of spectral clustering with those of Riemannian geometry \cite{allaoui2020considerably}. Its two key hyperparameters are the dimension of the output space $n_C$ (the number of components) and the number of neighbors $n_N$, which controls the balance between local and global structure \cite{kobak2021initialization}.

We refer to this secondary embedding as the refined embedding, since the purpose of the second stage is to further refine the latent representations to emphasize the global patterns in the embedded data thus making it more cluster-friendly. By applying UMAP after the compressed embedding we have the advantage of choosing DNN architecture based on the type of the data set. For example, for image data, a CAE presents the best choice. Therefore, interchanging the order of the first two stages is undesirable, since performing UMAP first leads to a loss of data structure that can be otherwise exploited with a particular choice of DNN architecture.

\textbf{Stage III: Clustering}

After obtaining the compressed (Stage I) and refined (Stage II) embeddings of the data, any conventional clustering algorithm can be applied to categorize the embedded data. Note that the choice of a particular clustering algorithm is not required to be correlated with the clusterability measure used to train the DNN auto-encoder in Stage I. For example, choosing in Stage I as a clustering loss the cost function of K-means may be followed by DBSCAN in Stage III. This flexibility is made possible by the secondary DR in Stage II. Specifically, UMAP improves the overall clusterability of the refined embedding not just with respect to a single clusterability measure trained for in Stage I, but most likely with respect to other such measures as well.

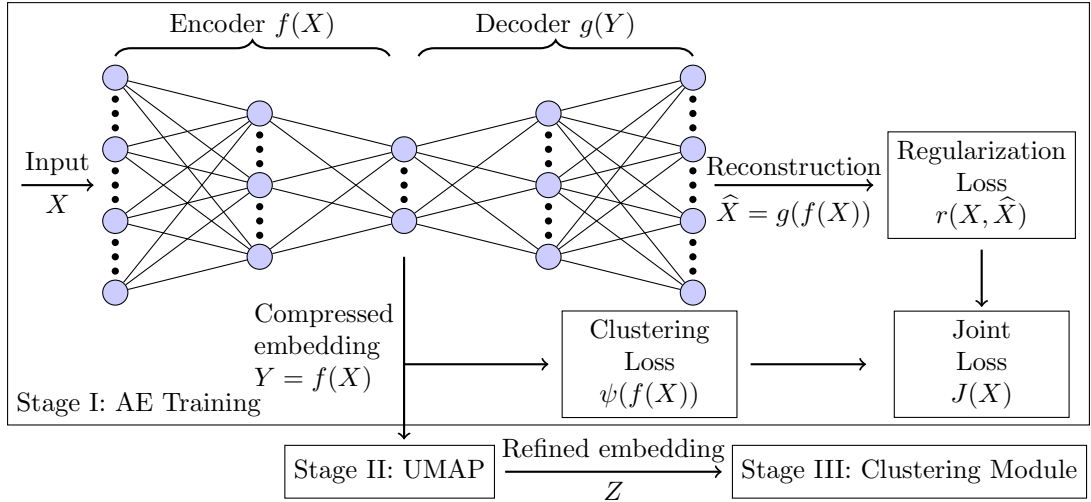
\begin{figure}
\centering
\begin{tikzpicture}[scale=0.95,node distance=0.75cm,
vertex/.style={circle, draw, fill=blue!20, minimum size=0.01cm, font=\tiny},
]
\node[draw, rectangle, text width=14cm, minimum width=10cm, minimum height=5.6cm] (mainNode) at (6,1.6) {};
\node[align=right, anchor=south west] at (mainNode.south west) {Stage I: AE Training};

\draw[->, thick] (-1.3,2) -- (-0.3,2) node[midway, above] {Input} node[midway, below] {$X$};

\foreach \i in {1,2,3,4} {
    \node[vertex] (v0-\i) at (0,\i-0.5) {}; 
}
\foreach \i in {0.8,1,1.2, 1.8, 2, 2.2, 2.8, 3, 3.2} {
    \fill[black] (0,\i) circle (0.05cm);
}
\foreach \i in {1,2,3} {
    \node[vertex] (v1-\i) at (2,\i) {};
}
\foreach \i in {1.3, 1.5, 1.7, 2.3, 2.5, 2.7} {
    \fill[black] (2,\i) circle (0.05cm);
}

\foreach \i in {1,2} {
    \node[vertex] (v2-\i) at (4,\i+0.5) {};
}
\foreach \i in { 1.8, 2, 2.2} {
    \fill[black] (4,\i) circle (0.05cm);
}
\foreach \i in {1,2,3} {
    \node[vertex] (v3-\i) at (6,\i) {};
}
\foreach \i in {1.3, 1.5, 1.7, 2.3, 2.5, 2.7} {
    \fill[black] (6,\i) circle (0.05cm);
    }
\foreach \i in {1,2,3,4} {
    \node[vertex] (v4-\i) at (8,\i-0.5) {}; 
}
\foreach \i in {0.8,1,1.2, 1.8, 2, 2.2, 2.8, 3, 3.2} {
    \fill[black] (8,\i) circle (0.05cm);
}
\foreach \i in {1,2,3,4} {
    \foreach \j in {1,2,3} {
        \draw (v0-\i) -- (v1-\j);
    }
}

\foreach \i in {1,2,3} {
    \foreach \j in {1,2} {
        \draw (v1-\i) -- (v2-\j);
    }
}

\foreach \i in {1,2} {
    \foreach \j in {1,2,3} {
        \draw (v2-\i) -- (v3-\j);
    }
}

\foreach \i in {1,2,3} {
    \foreach \j in {1,2,3,4} {
        \draw (v3-\i) -- (v4-\j);
    }
}

\draw[->, thick] (8.3,2) -- (10.5,2) node[midway, above] {Reconstruction} node[midway, below] {$\widehat{X}=g(f(X))$} ;
\node[draw, rectangle,align=center, minimum width=2.3cm, minimum height=1cm, fill=white] at (12.0,2) {Regularization\\Loss\\$r(X, \widehat{X}$)};

\draw[->, thick] (12.0,1.1) -- (12.0,0.3) ;

\draw[->, thick] (4,1.0) -- (4,-1.55) node[midway,left,align=left]{Compressed\\embedding\\$Y=f(X)$};

\draw[->, thick] (4,-0.5) -- (6,-0.5) node[midway,left] {};
\draw[->, thick] (8.8,-0.5) -- (10.4,-0.5) node[midway,left] {};

\node[draw, rectangle,align=center, minimum width=2.3cm, minimum height=1cm, fill=white] at (7.4,-0.5) {Clustering\\Loss\\$\psi(f(X))$};

\node[draw, rectangle,align=center, minimum width=2.3cm, minimum height=1cm, fill=white] at (12.0,-0.5) {Joint\\Loss\\$J(X)$};

\node[draw, rectangle, minimum width=2.5cm, minimum height=0.7cm, fill=white] at (3.8,-2.0) {Stage II: UMAP};

\draw[->, thick] (5.4,-2.0) -- (8.4,-2.0) node[midway, above] {Refined embedding} node[midway, below] {$Z$} ;

\node[draw, rectangle, minimum width=2.5cm, minimum height=0.7cm, fill=white] at (11.0,-2.0) {Stage III: Clustering Module};

\draw[decorate, decoration={brace, amplitude=6pt}, thick] (0,3.8) -- (3.8,3.8) node[midway, above=3pt] {Encoder $f(X)$};
\draw[decorate, decoration={brace, amplitude=6pt}, thick] (4.2,3.8) -- (8,3.8) node[midway, above=3pt] {Decoder $g(Y)$};
 
\end{tikzpicture}
\caption{Data flow in AUEC.}
\label{fig:data_flow}
\end{figure}

We summarize the three-stage AUEC framework in Algorithm~\ref{algo:auec}
with the corresponding data flow displayed in Figure~\ref{fig:data_flow}.
A Python implementation of AUEC is available at
\url{https://github.com/mchavooshi/AUEC}

\section{Compressed embedding computation}
\label{sec:compress}

We observe in Algorithm~\ref{algo:auec} that Stages II and III
of AUEC consist of a straightforward application of standard 
algorithms to the compressed and refined embeddings of the data, respectively. On the other hand, computing the compressed 
embedding in Stage I is more involved and represents the bulk 
of the technical content of AUEC. In this section we discuss 
the computation of the compressed embedding in detail.

\begin{algorithm}
\caption{AUEC}
\label{algo:auec}
\begin{algorithmic}[1]
\State \textbf{Input:} data set 
\( X = \{ x_i \}_{i=1}^{N} \subset \mathbb{R}^M \), the desired number of clusters \( K \), UMAP hyper-parameters: number of components \( n_C \) and number of neighbors \( n_N \).
\State \textbf{Stage I:} Train the autoencoder \( \widehat{x}_i = g(f(x_i; w_f); w_g) \), where 
\begin{equation} 
y_i = f(x_i; w_f), \quad i = 1,\ldots,N,
\label{eqn:yi}
\end{equation}
is the encoder with weights \( w_f \), and \( g(y_i; w_g) \) is the decoder with weights \( w_g \) and 
\( Y = \{ y_i \}_{i=1}^{N} \in \mathbb{R}^m \) is the compressed embedding of the data with $m \ll M$. The training is performed by solving the optimization problem
\[ \mathop{\mbox{minimize}}\limits_{w_f, w_g} J(X, K; w_f, w_g) \]
where \( J \) is the joint loss of the form
\begin{equation}
\begin{split}
J(X, K; w_f, w_g) = \lambda 
\psi\Big(f(X; w_f), K\Big) \\+ 
\rho \Big(X, g(f(X; w_f); w_g)\Big).
\end{split}
\label{eq:jointloss}
\end{equation}
Here \( \psi \) is the clustering loss, \( \rho \) is the regularization loss and \( \lambda>0 \) is the regularization parameter.
Once the AE training is complete, compute the compressed embedding \( Y \) via \eqref{eqn:yi} and pass it onto Stage II.
\State \textbf{Stage II:} Apply UMAP with hyper-parameters \( n_C \) and \( n_N \) to \( Y \) to obtain the refined embedding 
\( Z = \{ z_i \}_{i=1}^{N} \subset \mathbb{R}^{n_C} \), with
$n_C < m$.
\State \textbf{Stage III:} Apply a clustering algorithm of choice to \( Z \) to obtain the clustering \( C(Z) = \{C_1, \ldots, C_K \} \) of \( Z \), where $C_j \subseteq Z$, $j = 1,\ldots,K$, with
\begin{equation}
Z = \mathop{\sqcup}\limits_{j=1}^{K} C_j.
\end{equation}
\end{algorithmic}
\end{algorithm}

\subsection{Autoencoder and regularization loss}

\begin{figure}[htbp]
    \centering
    \includegraphics[width=0.8\textwidth]{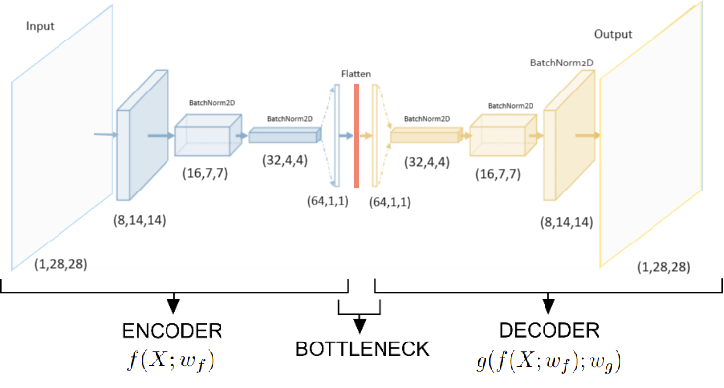} 
    \caption{The architecture of the CAE used to work with image data. Convolutional layers are displayed in blue, while transposed convolutional layers are shown in yellow. The red layer in the middle represents the bottleneck, which flattens the data for latent space representation. Batch normalization is applied after each layer marked with ``BatchNorm2D'', and ReLU activation is used after each layer.}
    \label{fig:CAE_structure}
\end{figure}

The main part of AUEC Stage I is training the autoencoder \( g(f(X; w_f); w_g) \). Thus, the key step is to choose its architecture. As mentioned above, this choice should exploit fully the structure of the data. In this paper we illustrate the performance of AUEC with an example of a dataset of images,
therefore it makes sense to employ a convolutional autoencoder (CAE). Specifically, for MNIST dataset one may choose a CAE architecture as in Figure~\ref{fig:CAE_structure}. 
The CAE is trained using the joint loss \eqref{eq:jointloss}
which requires defining both its components. We defer the discussion of the clustering loss to Section~\ref{sec:closs},
while we choose for the regularization loss the most commonly used reconstruction loss in autoencoder training, the Mean Squared Error (MSE). 
The MSE loss measures the dissimilarity between the CAE output and the input, quantifying the reconstruction error as
\begin{equation}
\text{MSE}(X, \widehat{X}) = \frac{1}{N} \sum_{i=1}^{N} \| x_i - \widehat{x}_i \|_2^2.
\label{eq:mse}
\end{equation}
Therefore, the regularization loss is defined by
\begin{equation}
\rho(X, g(f(X; w_f); w_g)) = 
\text{MSE} \Big(X, g(f(X; w_f); w_g) \Big).
\label{eq:reg_loss}
\end{equation}

\subsection{Clustering loss}
\label{sec:closs}

The main purpose of Stage I of AUEC is to produce a compressed embedding that is more clusterizable than the original data. This is achieved by the appropriate choice of the clustering loss in \eqref{eq:jointloss}. Thus, we should employ as a clustering loss a function that promotes clusters with intra-cluster points being in close proximity while ensuring considerable distance between points in different clusters. 
Such functions are sometimes referred to as clusterability measures. The simplest and the most widely used such measure is the so-called within clusters sum of squares (WCSS) which is the loss function of K-means defined as
\begin{equation}
\text{WCSS}(Y, C(Y)) = 
\sum_{j=1}^{K} \sum_{y_i \in C_j} \Big\| 
y_i - \frac{1}{|C_j|} \sum_{y_k \in C_j} y_k                \Big\|_2^2,
\label{eqn:wcss}
\end{equation}
where $C(Y) = \{ C_1, \ldots, C_K \}$ is the clustering of data $Y$. WCSS clusterability measure has been used before in the context of network training for unsupervised learning in \cite{li2018discriminatively,yang2017towards}. 
When used as a loss function for network training it has a disadvantage of requiring the data to be clustered at each training iteration, since one of the inputs of WCSS is the clustering $C(Y)$. 
This requires modifying the standard training procedures by splitting them into a part that updates the clustering assignment $C(Y)$ which is then frozen and passed to the part that updates the network weights.
In order to avoid these complications, we employ here as a clustering loss an alternative clusterabilty measure 
that is built using the spectral graph theory. 

The central concepts of the spectral graph theory \cite{chung1997spectral} 
are the graph-Laplacian matrix and its spectrum defined as follows. First, given the data \( Y \), one constructs a similarity matrix \( \mathbf{S} \in \mathbb{R}^{N \times N}\) where each entry \( s_{ij} \) measures the similarity between \( y_i \) and \( y_j \), while also setting \( s_{ii} = 0 \), e.g.,
\begin{equation}
s_{ij} = e^{- \gamma \| y_i - y_j \|^2 }, \quad \gamma > 0. 
\end{equation}
Then, introducing the diagonal degree matrix \( \mathbf{D}  \in \mathbb{R}^{N \times N} \) with entries
\( d_{ii} = \sum_{j=1}^{N} s_{ij} \) one defines the normalized 
graph-Laplacian as 
\begin{equation}
\mathbf{L} = \mathbf{I}_N - 
\mathbf{D}^{-\frac{1}{2}} \mathbf{S} \mathbf{D}^{-\frac{1}{2}},
\end{equation}
where $\mathbf{I}_N$ is the identity matrix. Next, consider the eigenvalues of $\mathbf{L}$ arranged in the non-decreasing order
\begin{equation}
0 = \lambda_1 \leq \lambda_2 \leq \ldots
\end{equation}
Classically, the spectral gap is the difference between its second smallest eigenvalue \( \lambda_2 \) and its smallest eigenvalue \( \lambda_1 = 0 \):
\begin{equation}
\Delta \lambda_1 = \lambda_2 - \lambda_1 = \lambda_2.
\end{equation}
A larger spectral gap signifies that the graph is well-connected.
Conversely, a smaller or near-zero spectral gap suggests the presence of weak connections or potential bottlenecks in the graph. In the extreme case $\gamma = \lambda_2 = 0$ the graph consists of two disconnected components.

When working with more than two clusters, one needs to generalize the spectral 
gap notion. This is motivated by the following spectral gap heuristic often used 
to decide how many clusters there are in the data \cite{von2007tutorial}. 
If the data admits $K$ well-defined clusters, one would expect the first $K$
eigenvalues to be small, while $\lambda_{K+1}$ should be relatively large. 
Similarly to the two clusters case, if the graph has $K$ disconnected components, 
the zero eigenvalue appears $K$ times with a noticeable gap before the next
eigenvalue $\lambda_{k+1} > 0$. In general, the first eigenvalues 
of $\mathbf{L}$ are related to the graph's topological properties 
like the sizes of graph cuts, see, e.g., \cite{chung1997spectral}.

Given the above, when classifying the data into $K$ clusters, it is useful to consider consecutive eigenvalues of $\mathbf{L}$, 
\( \lambda_{K+1} \) and \( \lambda_K \) and their differences
\begin{equation}
\Delta \lambda_K = \lambda_{K+1} - \lambda_K.
\end{equation}
Since we are interested in constructing a loss function out of the spectral gap, it is more convenient for the purposes of optimization to work with the relative quantity
\begin{equation}
\text{RSG}(Y, K) = \frac{\lambda_{K+1}}{\lambda_K}
= \frac{\Delta \lambda_K}{\lambda_K} + 1,
\label{eqn:rsg}
\end{equation}
that we refer to as the relative spectral gap (RSG). Note that computing \eqref{eqn:rsg} does not require the data to be clustered. Maximizing this quantity has an enhancing effect on clustering the data into $K$ clusters. 
The mechanism of this behavior is as follows. Increasing $\text{RSG}(Y,K)$ has the effect of pushing $\lambda_K$ down towards zero, while not letting $\lambda_{K+1}$ approach zero. In the best-case scenario, this will result into 
\begin{equation}
\lambda_1 = \ldots = \lambda_K = 0, \text{ while }
\lambda_{K+1} > 0,
\end{equation}
meaning that the graph is split into $K$ disconnected components, hence the data $Y$ is perfectly clustered into $K$ clusters. Therefore, $\text{RSG}$ can be used as a clusterability measure alternative to the conventional WCSS \eqref{eqn:wcss} with the higher values of RSG corresponding to better clustering results. This makes it a good candidate for a
clustering loss that can be defined as
\begin{equation}
\psi(Y, K) = \frac{1}{\text{RSG}(Y, K)}
\label{eq:rsgloss}
\end{equation}
where $Y = f(X; w_f)$. 

\subsection{Autoencoder training}

Once both the clustering loss and the regularization loss
components of the joint loss \eqref{eq:jointloss} are fixed,
the autoencoder can be trained using any off-the-shelf optimizer.
The only part of AE training that requires special attention is 
the initial guess for the weights $[w_f; w_g]$.
Given that the joint loss \eqref{eq:jointloss} is expressed as a 
sum of two terms, it makes sense to employ a simple pre-training
technique consisting of training the AE network
with just a regularization loss, i.e., setting $\lambda = 0$ in 
\eqref{eq:jointloss} for a small number of epochs 
(e.g., we pre-train the AE for 5 epochs in the numerical
experiments below). 

\section{Numerical experiments}
\label{sec:numerics}

In this section, we validate AUEC framework on MNIST dataset \cite{726791}. 
The dataset consists of grayscale images, each depicting a handwritten digit. 
The size of each image is $28 \times 28 = 784$ pixels. There are a total of $10$ image classes corresponding 
to the numerical digits (0 to 9). The dataset is labeled, i.e., for each digit its ground truth class is known. 
Even though AUEC is an unsupervised learning framework, we use the available ground truth 
labels to measure the accuracy of the clustering it produces.
The MNIST dataset consists of two subsets: a training set with 60,000 samples and a testing set with 10,000 samples. 
For most of our numerical experiments, we use the training set of $N = 60,000$ samples, while in Section~\ref{sec:robust} we use the testing set to assess the performance of AUEC on unseen data without retraining the AE in Stage I.

\subsection{Baseline Methods}
\label{sec:baseline}

We compare the performance of AUEC against a number of basic
unsupervised learning approaches, as well as a few leading 
state-of-the-art techniques:
\begin{enumerate}[\hspace{0pt}(1)]
    \item \textbf{KMS} is the baseline method that applies the K-means algorithm directly to raw image data. 

    \item \textbf{UMAP+KMS} utilizes UMAP to embed the images into a lower-dimensional space, 
    after which K-means clustering is performed on the UMAP-embedded data.

    \item \textbf{Deep Embedded Clustering (DEC)} of \cite{xie2016} learns feature representations and determines cluster centers using deep AE and soft K-means, respectively. Similarly to Stage I of AUEC, the deep AE model is trained with a joint loss with the clustering loss taken as the Kullback–Leibler (KL) divergence. 

    \item \textbf{Deep Clustering Network (DCN)} of \cite{yang2017towards} combines K-means clustering with 
    DR via deep SAE similar to DEC 
    However, instead of KL divergence of DEC, DCN utilizes the WCSS loss as the clustering component of its joint loss.
    
    \item \textbf{Fully Convolutional Autoencoder (FCAE)-KMS} of \cite{li2018discriminatively} is similar to DCN,
    but it adopts FCAE for image feature extraction.
\end{enumerate}

\subsection{Results}

The results of numerical experiments are presented below both numerically and visually. Numerically, we judge the performance of AUEC and other approaches via the three evaluation metrics discussed in Section~\ref{sec:compare}.
For visual evaluation of results, we rely on 2D embedding techniques, where appropriate. 
For the approaches involving UMAP (including AUEC itself), we rely on the embeddings it provides for $n_C = 2$. In all the 2D embeddings, we color-code the data points based on either their ground truth class labels or predicted cluster labels. This indicates the areas of agreement and disagreement between the ground truth classes and the predicted clusters.

\subsubsection{UMAP+KMS}
\label{sec:umapkms}

\begin{figure}[htbp]
    \centering
    \begin{subfigure}[t]{0.4\textwidth}
        \includegraphics[width=\textwidth]{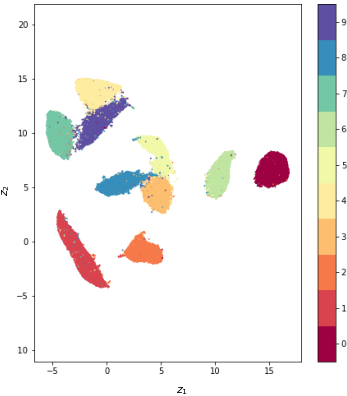}
        \caption{Ground truth labels.}

    \end{subfigure}
    \begin{subfigure}[t]{0.415\textwidth}
        \includegraphics[width=\textwidth]{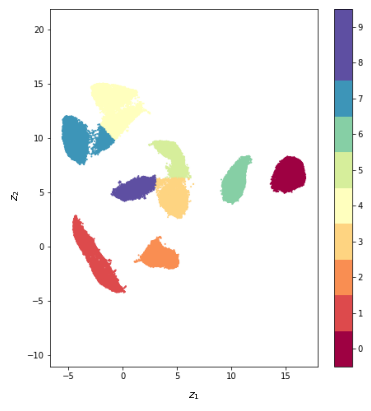}
        \caption{K-means predicted labels.}

    \end{subfigure}
\caption[UMAP-embedded MNIST data.]{UMAP-embedded MNIST data.}
\label{fig:umap+kms}
\end{figure}

Since UMAP is an integral part of the AUEC framework, it makes sense to study how it performs in a clustering setting while being the only DR transform applied to the data. 
We set UMAP hyper-parameters to $n_N = 15$ and $n_C = 2$ which corresponds to embedding the data into 
the 2D space for ease of visualization. After UMAP embedding we perform K-means and display the results in
Figure~\ref{fig:umap+kms}. 
We observe that the digits ``0'', ``1'', ``2'' and ``6'' are well-separated and almost perfectly recovered by K-means from the embedded data. However, there are two three-digit clouds that are harder to classify. These are the ``3-5-8'' and ``4-7-9'' clouds. We observe in the predicted labels plot that while K-means did a decent job separating the ``3-5-8'' cloud into three clusters, it failed to do so with the ``4-7-9'' cloud. In fact, it was only able to identify two clusters in that cloud instead of three and therefore only found 9 clusters in the UMAP-embedded data instead of 10.

\subsubsection{AUEC-MDBCAN}
\label{sec:auecrsg}

We present here the results of applying AUEC to MNIST data with a special choice of clustering algorithm for Stage III. In particular, we utilize a modification of DBSCAN \cite{ester1996density} in which
we enforce the output of $K = 10$ clusters by merging the smaller clusters and outliers that DBSCAN produces with the $10$ largest ones based on proximity. We refer to this variant of the framework as AUEC-MDBSCAN.

\begin{figure}[htbp]
    \centering
    \begin{subfigure}[t]{0.405\textwidth}
        \includegraphics[width=\textwidth]{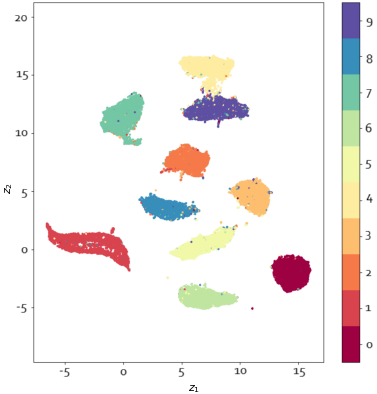}
        \caption{Ground truth labels.}
    \end{subfigure}
    \begin{subfigure}[t]{0.4\textwidth}
        \includegraphics[width=\textwidth]{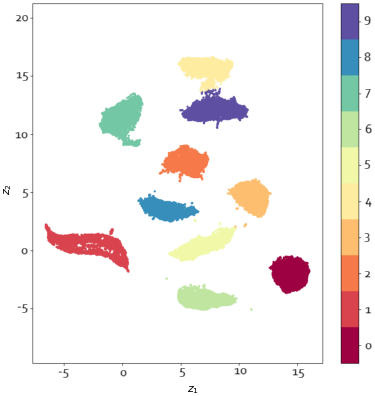}
        \caption{AUEC-MDBSCAN labels.}
    \end{subfigure}
    \caption{AUEC-MDBSCAN refined embedding of MNIST data.}
    \label{fig:auec-mdbscan}
\end{figure}

Similarly to the previous section, we display in
Figure~\ref{fig:auec-mdbscan} the refined embedding $Z$ of MNIST data computed by AUEC-MDBSCAN for $n_C = 2$ and $n_N = 5$.
Comparing the results to those in Figure~\ref{fig:umap+kms}
we observe an excellent separation of clusters. In particular, clusters ``3'' and ``5'' are now completely detached. All other digit classes are well-separated as well, except for a relatively thin neck connecting ``4'' and ``9''.

\begin{figure}[htbp]
    \centering
    \includegraphics[width=0.6\textwidth]{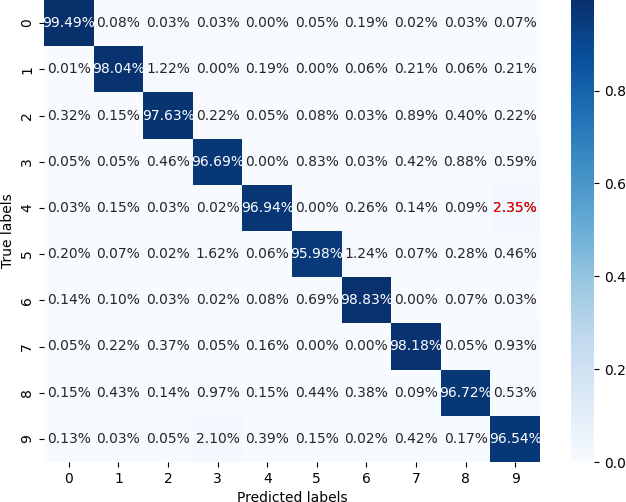}
    \caption{AUEC-MDBSCAN confusion matrix with the worst confusion shown in red.}
    \label{fig:confusion_matrix_rsg}
\end{figure}

\begin{figure}[htbp]
    \centering
    \includegraphics[width=0.9\textwidth]{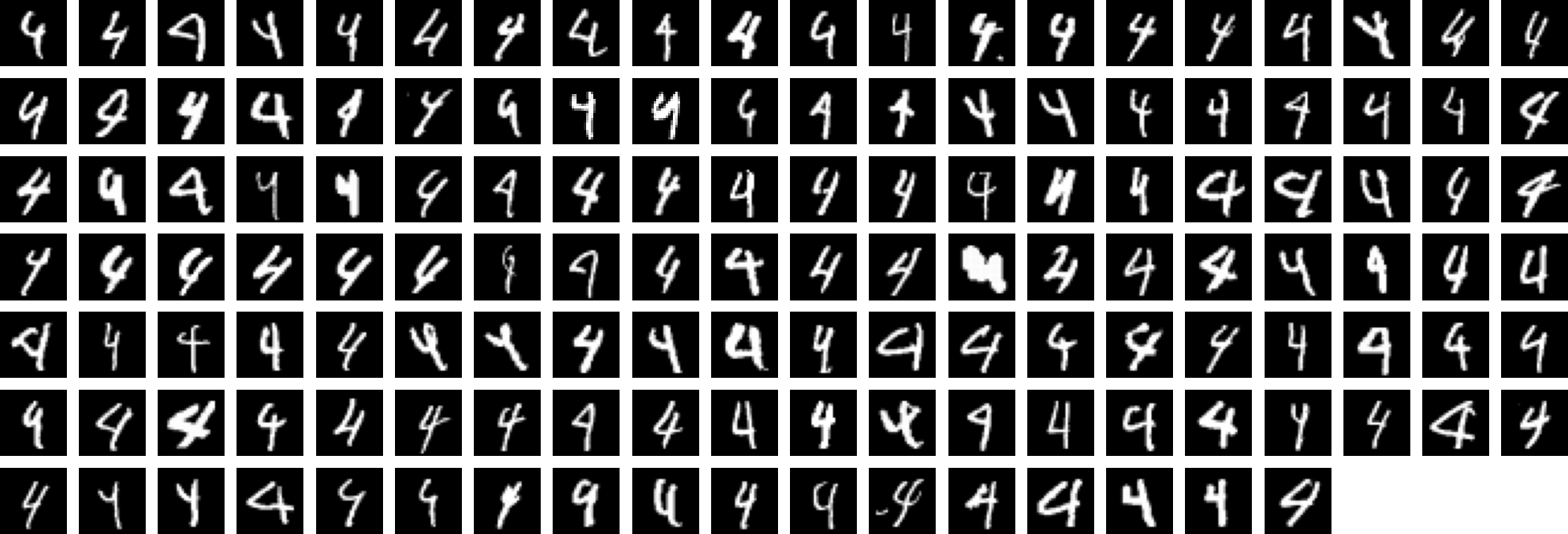}
    \caption{Digits corresponding to AUEC-MDBCSAN worst misclassification.}
    \label{fig:MisClassImages_rsg}
\end{figure}

To provide a deeper insight into the performance of AUEC-MDBSCAN we provide additional visualization in Figures~\ref{fig:confusion_matrix_rsg} and \ref{fig:MisClassImages_rsg}. These visualizations no longer require 
$n_C = 2$, so in order to achieve the best possible performance we set $n_C = 8$ with $n_N = 6$. 
In particular, in Figure~\ref{fig:confusion_matrix_rsg} we display the confusion matrix showing the percentage of digits in the intersections of each ground truth and predicted classes. Ideally, the confusion matrix should be diagonal with $100\%$ main diagonal entries and zero entries elsewhere. 
Any misclassified digits manifest themselves as non-zero off-diagonal entries with the largest such entry referred 
to as the worst misclassification. 
As expected from the results in Figure~\ref{fig:auec-mdbscan}, we observe that the worst misclassification occurs for the classes ``4'' and ``9'' with $2.35\%$ of ground truth ``4'' digits misclassified as ``9''. A total of $137$ such digits are displayed in Figure~\ref{fig:MisClassImages_rsg}. Indeed, to a human eye many of these digits look like ``9''.
Visually, AUEC-MDBSCAN is vastly superior to UMAP+KMS which is also confirmed numerically in the next section.

\subsection{Comparative Analysis}
\label{sec:compare}

Here we compare AUEC to the other approaches listed in 
Section~\ref{sec:baseline} using the standard metrics for assessing clustering performance. Specifically, we employ the three key metrics: normalized mutual information (NMI) \cite{cai2010locally}, adjusted Rand index (ARI) \cite{yeung2001}, and clustering accuracy (ACC) \cite{cai2010locally}. 
We summarize the testing results of Sections~\ref{sec:umapkms}--\ref{sec:auecrsg} in Table~\ref{tab:clustering_methods_comparison} while also comparing them to the results for DEC, 
DCN, 
and FCAE-KMS. 
The results for AUEC-MDBSCAN shown in Table~\ref{tab:clustering_methods_comparison} correspond to 
$n_C = 8$ with $n_N = 6$.

\begin{table}
\caption{Comparative performance of unsupervised learning methods on MNIST data.}
\centering
\begin{tabular}{|c|c|c|c|}
\hline
Methods & ACC & NMI & ARI \\
\hline
KMS & 59.07\% & 50.95\% & 40.47\% \\
UMAP+KMS & 86.59\% & 85.73\% & 80.41\% \\
DEC & 84.30\% & - & - \\
DCN & 83\% & 81\% & 75\% \\
FCAE-KMS & 79.4\% & 69.8\% & - \\
\textbf{AUEC-MDBSCAN} & \textbf{97.52\%} & \textbf{93.46\%} & \textbf{94.64\%} \\
\hline
\end{tabular}
\label{tab:clustering_methods_comparison}
\end{table}

We observe that AUEC vastly outperforms the existing approaches achieving an impressive accuracy of $97.5\%$.
This is due to a powerful combination of AE and UMAP that AUEC employs. Indeed, UMAP by itself is 
such a powerful DR technique that simply combined with K-means it outperforms the DNN-based 
approaches from \cite{xie2016, yang2017towards, li2018discriminatively}. Combining UMAP with an 
AE network trained with a carefully chosen clustering-promoting loss results in an even more powerful 
approach that is AUEC. In fact, when applied to MNIST dataset, AUEC-MDBSCAN scores fourth 
(as of August 2024) in terms of ACC in ``Unsupervised Image Classification on MNIST'' according to
\begin{sloppypar}
\noindent
\url{https://paperswithcode.com/sota/unsupervised-image-classification-on-mnist}
\end{sloppypar}

\subsection{Robustness study}
\label{sec:robust}

We conclude the numerical experiments with a robustness study of 
AUEC-MDBSCAN. For this study we perform Stage I of AUEC-MDBSCAN on MNIST training data. 
Then, we perform Stages II and III on MNIST test data and compute the evaluation metrics. 
Even though we expect a certain decrease in ACC, NMI and ARI since the compressed embedding 
is not fine-tuned for the previously unseen data, we still anticipate a relatively high level of accuracy if 
Stage I of AUEC is robust. Our expectations are confirmed by Table~\ref{tab:robust}.

While we lose about $2\%$ of ACC and $3-4\%$ of NMI and ARI, the evaluation metrics are still quite high, 
which means that the AUEC framework is robust without retraining Stage I. 
Note that Stage I is the most computationally expensive part of AUEC. 
Thus, in practice, one may perform Stage I on a smaller representative dataset and use the trained AE 
to apply AUEC to a much larger dataset without significant loss of accuracy.

\begin{figure}[htbp]
    \centering
    \begin{subfigure}[t]{0.4\textwidth}
        \includegraphics[width=\textwidth]{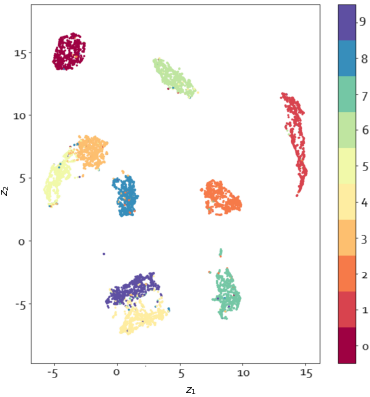}
        \caption{Ground truth labels.}
    \end{subfigure}
    \begin{subfigure}[t]{0.4\textwidth}
        \includegraphics[width=\textwidth]{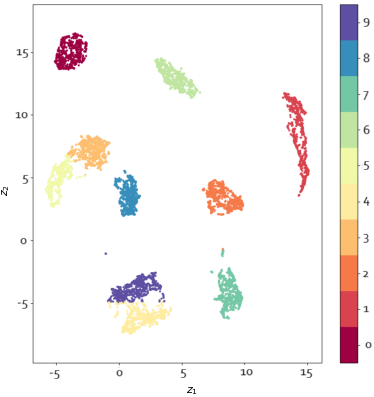}
        \caption{AUEC-MDBSCAN labels.}
    \end{subfigure}
    \caption{AUEC-MDBSCAN refined embedding of MNIST test data. }
    \label{fig:robust1}
\end{figure}

In addition to reporting the results in table form, we display in
Figure~\ref{fig:robust1} the AUEC-MDBSCAN refined embedding of MNIST test data. 
Similarly to Figure~\ref{fig:auec-mdbscan} the clusters are well-separated, but the ``3'' and ``5'' clusters touch, 
which is a likely source of the loss of accuracy reported in Table~\ref{tab:robust}.

\begin{table}
\centering
\caption{Evaluation metrics for AUEC-MDBSCAN on MNIST training and test data without recomputing Stage I.}
\begin{tabular}{|l|c|c|c|}
\hline
MNIST data & ACC & NMI & ARI \\
\hline
Training & 97.52\% & 93.46\% & 94.64\% \\
\hline
Test & 95.57\% & 90.06\% & 90.60\% \\
\hline
\end{tabular}
\label{tab:robust}
\end{table}

\section{Conclusion}
\label{sec:conclude}

In this work, we introduced the AUEC framework for unsupervised learning.
The framework consists of three stages. First, an AE is trained to learn
a compressed embedding of the data using a carefully chosen regularized 
clustering-promoting loss function. The architecture of the AE 
is chosen based on the type of incoming data, e.g., one may employ CNN 
to handle image data.
Second, we apply UMAP to refine the data embedding, reduce further the 
data dimensionality, and make clustering easier. The presence of the embedding 
refinement stage makes the choice of algorithm for subsequent clustering more
flexible since it no longer has to match the clusterability criterion used to
construct the loss function in the first stage. 
Finally, a clustering algorithm of choice is applied to the refined embedding 
of the data, e.g., K-means or modified DBSCAN.

We applied AUEC framework to a classical ML example of clustering MNIST data. 
In this setting AUEC yields the accuracy (ACC) surpassing $97.5\%$, which compares favorably 
to the state-of-the-art unsupervised learning approaches and demonstrates the power of the framework. 
The framework is also robust as it loses only $2\%$ of accuracy when presented with 
unseen data without retraining the AE.

\backmatter

\bmhead{Acknowledgements}

A.M. and M.C. were supported by the U.S. National Science Foundation under awards DMS-1619821 and DMS-2309197.
This material is based upon research supported in part by the U.S. Office of Naval Research under award number 
N00014-21-1-2370 to A.M.

\bibliography{References}

\end{document}